\documentclass[sigconf]{acmart}

\AtBeginDocument{%
  }

\copyrightyear{2025} 
\acmYear{2025} 
\setcopyright{acmlicensed}\acmConference[KDD '25]{Proceedings of the 31st ACM SIGKDD Conference on Knowledge Discovery and Data Mining V.2}{August 3--7, 2025}{Toronto, ON, Canada}
\acmBooktitle{Proceedings of the 31st ACM SIGKDD Conference on Knowledge
 Discovery and Data Mining V.2 (KDD '25), August 3--7, 2025, Toronto, ON, Canada}
\acmDOI{10.1145/3711896.3736829}
\acmISBN{979-8-4007-1454-2/2025/08}

\newcommand{\eat}[1]{}
\newcommand{\modelname}{Citss}
\usepackage{algorithm, algorithmicx, algpseudocode}
\usepackage{balance}
\usepackage{tcolorbox}

\begin{document}

\title{Adapting Pretrained Language Models for Citation Classification via Self-Supervised Contrastive Learning}

\author{Tong Li}
\email{tlice@connect.ust.hk}
\orcid{0009-0000-6486-8528}
\affiliation{%
  \institution{Hong Kong University of Science and Technology}
  \department{Department of Computer Science and Engineering}
  \city{Kowloon}
  \state{Hong Kong SAR}
  \country{China}
}

\author{Jiachuan Wang}
\email{jwangey@connect.ust.hk}
\orcid{0000-0001-6473-8221}
\affiliation{%
  \institution{Hong Kong University of Science and Technology}
  \department{Department of Computer Science and Engineering}
  \city{Kowloon}
  \state{Hong Kong SAR}
  \country{China}
}

\author{Yongqi Zhang}
\authornote{Corresponding author.}
\orcid{0000-0003-2085-7418}
\email{yzhangee@connect.ust.hk}
\affiliation{
  \institution{Hong Kong University of Science and Technology (Guangzhou)}
\department{Thrust of Data Science And Analytics}
  \city{Guangzhou}
  \state{Guangdong}
  \country{China}
  }

\author{Shuangyin Li}
 \email{shuangyinli@scnu.edu.cn}
\orcid{0000-0001-6404-3438}
\affiliation{%
 \institution{South China Normal University}
 \department{School of Computer Science}
 \city{Guangzhou}
  \state{Guangdong}
  \country{China}}

\author{Lei Chen}
\email{leichen@ust.hk}
\orcid{0000-0002-8257-5806}
\affiliation{%
  \institution{Hong Kong University of Science and Technology}
  \department{Department of Computer Science and Engineering}
  \city{Kowloon}
  \state{Hong Kong SAR}
  \country{China}}


\begin{abstract}
Citation classification, which identifies the intention behind academic citations, is pivotal for scholarly analysis.
Previous works suggest fine-tuning pretrained language models (PLMs) on citation classification datasets, reaping the reward of the linguistic knowledge they gained during pretraining. 
However, directly fine-tuning for citation classification is challenging due to labeled data scarcity, contextual noise, and spurious keyphrase correlations.
In this paper, we present a novel framework, \modelname, that adapts the PLMs to overcome these challenges. 
\modelname~ introduces self-supervised contrastive learning to alleviate data scarcity, and is equipped with two specialized strategies to obtain the contrastive pairs: sentence-level cropping, which enhances focus on target citations within long contexts, and keyphrase perturbation, which mitigates reliance on specific keyphrases. 
Compared with previous works that are only designed for encoder-based PLMs, \modelname~ is carefully developed to be compatible with both encoder-based PLMs and decoder-based LLMs, to embrace the benefits of enlarged pretraining.
Experiments with three benchmark datasets with both encoder-based PLMs and decoder-based LLMs demonstrate our superiority compared to the previous state of the art. Our code is available at: github.com/LITONG99/Citss
\end{abstract}

\ccsdesc[500]{Computing methodologies~Information extraction}

\keywords{Citation Classification, Pretrained Language Models, Contrastive Learning}


\maketitle

\section{Introduction}
In scholarly writings, citations act as intellectual bridges, linking researchers and their ideas across time and disciplines to provide a connected view of the scientific literature. 
The analytic study of citations is the cornerstone for understanding the structure, evolution, and impact of scientific contributions, attracting growing attention in recent years~\cite{sciteai, kunnath2021meta}.
One of the critical focuses is the citation classification, which identifies and categorizes the authors' intention of using citations in their writing, empowering a range of applications, including research evaluation~\cite{valenzuela2015identifying, jurgens2018measuring, zhang2024pst}, research trends identification~\cite{valenzuela2015identifying, hassan2018novel}, paper recommendation~\cite{ren2014cluscite,sun2017recommendation, ding2022tell}, and scientific texts summarization~\cite{jha2017nlp, chen2019automatic, syed2023citance}. 

For a specific citation, its surrounding textual context is essential for revealing the underlying intention behind its explicit mentioning.
Traditional works in citation classification extract informative features from the contexts and then train supervised classifiers to assign labels~\cite{kunnath2021meta}. 
Some research~\cite {jurgens2018measuring} relies on hand-engineered features involving the in-text cue words, metadiscourse, part-of-speech tags, dependency relationships, etc. 
Others~\cite{cohan2019structural, roman2021citation} also bring in sophisticated deep-learned features such as the word representations from popular embedding models~\cite{pennington-etal-2014-glove, peters2018elmo, conneau2018infersent}. 
Nevertheless, these methods struggle to capture the semantic nuances between different citation intentions due to their limited model capacity.

Recent strides in transformer-based pretrained language models (PLMs) offer significant opportunities for developing more effective citation classification systems. 
Plenty of studies~\cite{kunnath2023prompting, shui2024fine} have demonstrated that PLMs acquired extensive linguistic knowledge during pretraining, which can be fine-tuned on citation classification datasets to discern subtle semantic nuances in citation intentions.
They leverage encoder-based PLMs, such as BERT~\cite{bert}, Longformer~\cite{beltagy2020longformer},
SciBERT~\cite{beltagy2019scibert}, to encode the context into representations, which is optimized end-to-end to automatically mine rich semantic information for the task.
Despite their efforts in fine-tuning PLMs on citation data, these methods are limited in addressing the following unique challenges and only consequent with sub-par performance.  


\textbf{Challenge 1: Scarcity of labeled citation data.}
It usually requires domain-specific knowledge and expertise of annotators to accurately interpret the scientific texts and assign citation labels. Hence, existing approaches either call on author self-annotation \cite{pride2019act, kunnath2022act2} or employ annotators with relevant academic backgrounds~\cite{grezes2023function, jurgens2018measuring}. To date, publicly available citation classification datasets remain limited to a few thousand samples, which fall short in manifesting the task-specific textual patterns and restrain the performance of deep learning systems~\cite{perier2019preliminary, kunnath2021meta}. With a great number of parameters, fine-tuning PLMs is more vulnerable confronted data scarcity.

\textbf{Challenge 2: Defocusing on the target citation.}
Most existing methods~\cite{kunnath2023prompting, maheshwari2021scibert, shui2024fine} only work with a highly related but extremely local context, which is the sentence directly containing the citation, referred to as the \textit{citance} ~\cite{syed2023citance, cohan2018scientific}.
However, the necessary semantic clues that enable us to determine the type of citation can be far from the citation and even fragmented in the writing~\cite{kunnath2022dynamic}.
Although the larger input window of PLMs makes it possible to include such long-range dependencies, a broader context can also introduce excessive irrelevant descriptions, such as mentions of other citations and general discourses, which are likely to distract the model from the target citation ~\cite{berrebbi2022graphcite} and even cause the problem of \textit{lost-in-the-middle}~\cite{liu2024lost}.

\textbf{Challenge 3: Spurious correlation based on keyphrases.}
The scientific keyphrases that are repeatedly mentioned in the texts, including research subjects, tasks, techniques, etc., are usually semantically significant for the PLMs, implying that the PLMs can easily establish spurious correlations from them to the label.
For example, in the case where each citation context in the training data discusses a unique technique, the model can easily fit on the dataset by learning a toxic mapping from the keyphrases to the observed label, resulting in a corrupted model. This problem can be further intensified by the insufficient training data discussed in Challenge 1.

In this paper, we address the above challenges and propose a framework, \modelname, that adapts pretrained language models for \textbf{Cit}ation classification via \textbf{s}elf-\textbf{s}upervised contrastive learning. 
For Challenge 1, our framework is equipped with two transform strategies, sentence-level cropping (SC) and keyphrase perturbation (KP), which generate contrastive pairs for citation classification in a self-supervised manner and derive contrastive loss to provide extra supervision signals for model fine-tuning, alleviating the demand for annotated citation contexts
Given the original sample, each of the SC and KP strategies not only produces diverse and realistic artificial samples serving as its contrastive pairs, but is also deliberately aimed to enhance the ability of PLMs in facing Challenge 2 and Challenge 3 correspondingly. 
Specifically, SC facilitates the model to focus on the target citation and improve model robustness against irrelevant noises, and KP helps to mitigate the spurious correlation established between specific keyphrases and the observed label. 

Besides developing our fine-tuning method specialized in elevating those previously highlighted encoder-based PLMs for citation classification, we are also ambitious to embrace the currently booming large language models (LLMs). 
With the number of parameters scaling up to billions, the advantages of pretraining are amplified for these LLMs~\cite{brown2020language} with decoder-based architecture,  making it appealing to harness their power for citation classification.
However, according to the latest attempts~\cite{nishikawa2024exploring, kunnath2023prompting}, adopting the cutting-edge LLMs, such as GPT-3.5-turbo, GPT-4, and SciGPT2~\cite{luu-etal-2021-explaining}, for citation classification in a language generation style still falls behind the fine-tuned "small" encoder-based PLMs, leaving the question of how to benefit from LLMs on citation classification open. 
In this regard, we establish our framework carefully so it can not only be applied to the decoder-based architectures of LLMs but also seamlessly incorporated with the prevalent parameter-efficient fine-tuning (PEFT) paradigms, such as Lora~\cite{yu2023low}, to further reduce the trainable parameters in LLMs. 
With our framework, we successfully fine-tuned a Llama3-8B backbone on the existing limited citation classification data and achieved noticeable improvements compared with baselines.       

To summarize, our main contributions are as follows
\begin{itemize}
\item We propose a novel self-supervised contrastive learning framework that is applicable to fine-tuning both encoder-based PLMs and decoder-based LLMs\footnote{We use ``PLMs'' to collectively refer to both types.} for citation classification under the scarcity of labeled citation data.
To the best of our knowledge, this is the first work to effectively fine-tune LLMs for citation classification. 

\item We propose a sentence-level cropping strategy that enhances the ability of PLMs to extract beneficial information for the target citation from long contexts and defend against irrelevant noises.

\item We propose a keyphrase perturbation strategy that assists the PLMs in predicting the citation intentions based on the context logic rather than the occurrence of specific keyphrases.

\item Experiments on three datasets with both an encoder-based PLM and a decoder-based LLM demonstrate the consistent superiority of our framework.
\end{itemize}

\section{Related Work}
The analytical study of citations boasts a long scholarly history. Within this domain, the citation classification task we investigate belongs to natural language processing grounded in citation context analysis~\cite{hernandez2016survey}. 
In the literature on related data mining research, this task has been specifically designated as either "citation function classification"~\cite{grezes2023function, kunnath2021meta, jurgens2018measuring} or "citation intent classification"~\cite{shui2024fine, roman2021citation, berrebbi2022graphcite,cohan2019structural, tsai2023citation}. While some scholarly works posit distinctions between these two concepts, in a sense that the former adopts an objective perspective centered on how citations serve scholarly writing, whereas the latter emphasizes authors' subjective psychological processes~\cite{nicolaisen2007citation}, we observe that their classification schemas are frequently identical or substantially overlapping. Consequently, this study refrains from further differentiation between these terminologies and collectively refers to them as "citation classification."

Other tasks that categorize the citations based on the contexts include: polarity classification~\cite{yu2020identifying}, which identifies authors' sentiment stance toward cited content; influence classification~\cite{zhang2024pst,kunnath2021overview}, which seeks to define the significance of a cited work and recognize the influential citations; role classification~\cite{zhao2019context}, which discerns the types of resources (code, data, website, media, etc.) provided by the citation link. These endeavors primarily serve specific application objectives, and their classification schemas exhibit substantial divergence from the research focus of this study, with marked differences in problem characteristics.

Additional citation analytical tasks may not necessarily originate from citation context analysis. For instance, citation prediction~\cite{haohlm, jin2023patton, zhang2023pre, cohan2020specter} aims to forecast potential citations from a candidate paper collection for the target document. Likewise, citation recommendation ~\cite{he2010context} attempts to recommend scholarly references to the authors during the paper drafting. 
These tasks differ more profoundly from the present research, primarily in that they focus on the stages where formal academic texts and actual citations may not yet exist.

\section{Preliminary}
\subsection{Citation Classification}
For the target citation $i$, its \textit{citation anchor}~\cite{ahmad2017pattern, ahmad2018cad} in the text indicates the occurrence of the citation, which can be in various formats (numerical, author-with-year, etc.) depending on the writing conventions. 
The citation context $T_i$ is the textual content surrounding the citation anchor~\cite{hernandez2016survey}. As shown in Figure~\ref{fig:framework} (b), we replace the citation anchor with a special tag (\#CITATION\_TAG) to distinguish it from other citations. 
Given the types of citation intentions $\mathcal{C}$, our task takes $T_i$ as the input and outputs the predicted label $y_i\in\mathcal{C}$.

\subsection{Pretrained Language Models}
From a high level, the interaction between our framework and the backbone PLM can be regarded as a function, 
\begin{equation}
    x=\mathcal{M}(T, \mathrm{\textbf{P}}),
    \label{eq:plm}
\end{equation} where $x$ is the output vector, $T$ is the citation context, and \textbf{P} is a textual prompt. 
\textbf{P} contains possible meta-information of the citation classification tasks, and defines the format in which $T$ is presented to the PLM, in order to help the model adapt to the specific task. 
The formatted input will be converted into a series of tokens and forwarded through the stacked transformer layers of the backbone $\mathcal{M}$. 
Correspondingly, there will be a series of hidden state vectors at the last layer, and we retrieve $x$ from them in different ways for encoder-based PLMs and decoder-based LLMs, depending on their distinct characteristics of language modeling.
(1) The \textbf{encoder-based PLM} consists of an encoder that is pre-trained with the blank infill task (masked language modeling). They are the leading models for numerous supervised text classification tasks~\cite{bucher2024fine}, among which the SciBERT~\cite{beltagy2019scibert} is optimized for scientific text and pertinent to citation classification.
Because these models are skilled at reconstructing information at the masked input token, we insert a mask token in the prompt and read out $x$ at the masked position.
(2) The \textbf{decoder-based LLM}, such as GPT-4, Llama3-8B, and Llama3-70B, achieves groundbreaking performance recently in various natural language understanding task~\cite{brown2020language}.
Unlike traditional classifiers, these models usually leverage language generation for text classification through question-answering or instruction-based paradigms~\cite{nishikawa2024exploring}. 
Architecturally, they employ decoder-only structures pretrained via next-token prediction objectives~\cite{chen2024next}.
Their output hidden state in the last position contains information for generating the next token, and is authentically intended to be decoded into a textual response. 
In our framework, instead of decoding, we incorporate the novel "Explicit-One-word-Limitation" trick (EOL)~\cite{jiang2023scaling}, which compels the model to condense all contextual understanding into the immediate next token's representation by explicitly instructing it to output exactly one more word in the prompt. 
We will then read $x$ out at the last position.

As the interaction with PLMs formulated as Equation~\ref{eq:plm}, our methodology only cares about how to prompt for the backbone model and to obtain the $x$ properly but is indifferent to the inner architectures. Therefore, our framework can work together with those PEFT methods~\cite{han2024parameter}, which select, reparameterize~\cite{yu2023low}, or insert~\cite{houlsby2019parameter} trainable modules in the PLMs to reduce trainable parameters while keeping the input and output interface unchanged, making it possible to efficiently fine-tune the LLMs with the limited citation data.

\subsection{Self-Supervised Contrastive Learning}
Contrastive learning~\cite{chen2020simple, oord2018representation} generates additional supervision signals by utilizing the similarity between data samples.
It first defines the \textit{contrastive pairs}, i.e., 
positive pairs and negative pairs, 
then the contrastive loss forces the positive pairs to be similar in the latent space, while the negative pairs to be dissimilar, encouraging the model to capture invariant information between positive pairs as well as distinguishable information between individual negative pair.
There are many ways to construct contrastive pairs.
In a self-supervised manner~\cite{wu2020clear}, carefully designed strategies can be employed to make modifications on the original sample $T_i$ and transform it into its positive pair $\tilde{T}_i$.
Other transformed samples in the batch, $\{\tilde{T}_{j\neq i}| j\in \mathcal{B}\}$, will serve as the negative pairs. 
With $z_i$ and $\tilde{z}_i$ denoting the representations of $T_i$ and $\tilde{T}_i$ in the latent space, the classic InfoNCE loss~\cite{oord2018representation} for the batch can be written as  
\begin{equation}
    L^{InfoNCE} = -\frac{1}{|\mathcal{B}|}\sum_{i\in\mathcal{B}} \log \frac{\exp{sim(z_i,\tilde{z}_i)}}{\sum_{j\in\mathcal{B}} \exp{sim(z_i,\tilde{z}_j)}},
\end{equation}
where $sim(q,k)=q^\mathsf{T}k/\tau$ is the similarity metric and $\tau$ is the temperature hyperparameter that adjusts the strength of contrast.

In order to benefit from the contrastive learning for citation classification, it is substantial to develop proper transformation strategies tailored to the task.
On the one hand, indicative information about the citation intention contained in the original sample is supposed to be preserved after the transformation. 
On the other hand, the transformation needs to introduce sufficient input diversity to effectively guide the model to gain discriminative ability in desired aspects. 

\begin{figure*}
    \centering
    \includegraphics[trim=0 200 0 0 , clip, width=\textwidth]{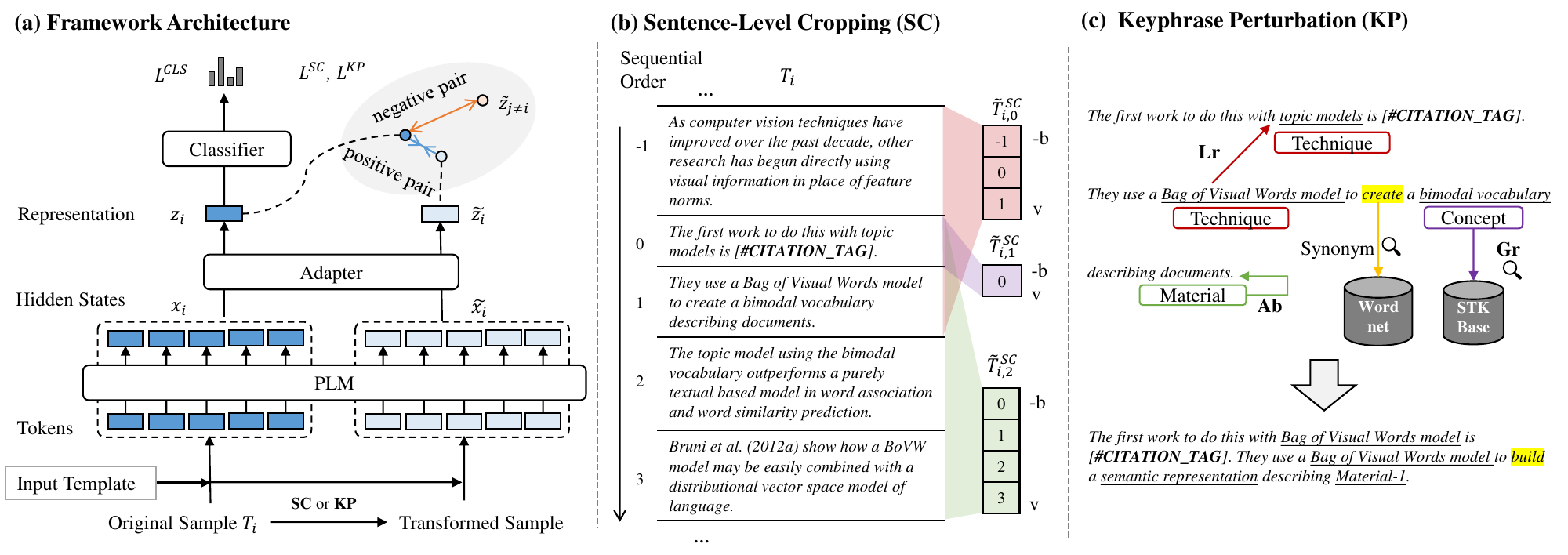}
    \caption{Overview of \modelname: (a) exhibits the architecture and workflow. (b) shows an example of sentence-level cropping. (c) shows an example of the keyphrase perturbation. The \underline{underlined text} is the keyphrases and the shaded text is the word for synonym replacement.}
    \label{fig:framework}
\end{figure*}
\section{Methodology}
\eat{
\subsection{Overview}
For the original sample, our framework first employs the sentence-level cropping and keyphrase perturbation strategies to produce the transformed samples.
Then, the original and transformed samples are wrapped in the prompt and separately sent into the PLM to retrieve the output vector from the final hidden states.
Next, an adapter module is applied to acquire task-specific context representation from the direct output of the PLM. 
In the latent space of the context representations, the original sample and its transformed samples are regarded as the positive pair to compute self-supervised contrastive loss $L^{SC}$, $L^{KP}$, creating two learning objectives without knowing the label.
Finally, the original representation is fed into the classifier for prediction, and calculate the classification loss $L^{CLS}$ over the labeled data. 
In the following methodology, we will first present the framework formally and discuss the training and inference. Then, we will elaborate on the two transformation strategies.}

\subsection{Framework Architecture}%
An overview of \modelname~ is depicted in 
Figure~\ref{fig:framework} (a). 
In this section, we will first describe the framework architecture to show how to obtain context representations and create extra supervision signals by contrastive learning.
Then, we will discuss the sentence-level cropping strategy and explain why it helps the model to focus on the target citation against contextual noises.
Next, we will elaborate on the definition and algorithm of the keyphrase perturbation strategy and how it helps to mitigate the spurious correlations.
Finally, we derived the complexity of our framework. 

Given the context $T_i$ of sample $i$, each of the two strategies, sentence-level cropping (SC) and keyphrase perturbation (KP), will modify it into the corresponding transformed sample $\tilde{T_i}$.
$T_i$ and $\tilde{T_i}$ are input into the PLM separately to obtain the hidden state vector, $x_i=\mathcal{M}(T_i, \mathrm{\textbf{P}}),$ and $\tilde{x_i}=\mathcal{M}(\tilde{T_i}, \mathrm{\textbf{P}}).$

The output vectors directly from the PLM comply with the intrinsic distribution of the hidden states, thereby limiting in characterizing task-specific information for citation classification. Therefore, we introduce a lightweight MLP adapter module to map them into task-specific context representation. 
The adapter is also able to perform dimension reduction, so that the subsequent modules can compute the similarity between contrastive pairs in an appropriate latent space. 
Formally, 
\begin{equation}
    f(x) = W_2\mathsf{LN}(\mathsf{GeLU}(W_1x+b_1))+b_2,\\
\end{equation}where $\mathrm{LN}(\cdot)$ is the layer normalization, $\mathrm{GeLU}(\cdot)$ is the activation function, and $W_1, W_2, b_1,b_2$ are learnable parameters. 
%
The representation of the original context $z_i=f(x_i)$ is sent to a linear classifier for prediction, and calculate the multi-class classification loss.
\begin{equation}
    y_i = g(z_i) = \mathsf{softmax}(W_3 z_i+b_3),    
\end{equation}
\begin{equation}
    L^{CLS}= \sum_{i\in\mathcal{B}} CrossEntropy(y_i, \hat{y}_i),
\end{equation} where $\mathcal{B}$ is the batch index set and $W_3, b_3$ are learnable parameters.

As for the representations of transformed contexts $\tilde{z_i}=f(\tilde{x_i})$, they are only used for contrastive learning. Under each strategy, we compute the the InfoNCE loss $L^{SC}$ or $L^{KP}$, so the overall optimization target is 
\begin{equation}
L= L^{CLS} +\lambda_1 L^{SC} + \lambda_2 L^{KP} + \omega L^{pnt},
\end{equation} where $\lambda_1,\lambda_2,\omega$ are hyperparameters controlling the magnitude of the loss terms, and $L^{pnt}$ is the weight decay penalty loss~\cite{loshchilov2017decoupled} for overfitting prevention.
It is worth mentioning that the transformations and contrastive learning are only conducted for the fine-tuning stage to aid parameter learning. During inference, we collect the citation context and forward it sequentially through modules for prediction.

\subsection{Sentence-Level Cropping}
The sentence-level cropping helps the model focus on the target citation with the presence of contextual noises, and its intuition resembles image cropping~\cite{devries2017improved, chen2020simple} in computer vision, which guides the model to attend on the target object against background noises.%
In the textual context, we crop on the sentence level because a sentence is a unit to convey a complete thought in natural language. 

As illustrated in Figure~\ref{fig:framework} (b), SC splits the long context into a sequence of sentences. In every epoch, it randomly crops a subsequence with the same citance but different context ranges. 
Minimizing the contrastive loss imposes the representation to be correlated with the target citation regardless of the randomly allocated input range, thus encouraging the model to focus on the target citation within long input. 
Rather than only containing the citance, each positive pair may also randomly overlap on a portion of surrounding sentences with the original sample, so the representation immediately after this optimization step tends to encode such random contextual information.
Supposing the information contributes to the correct prediction, it is likely to lead to better classification loss in the following steps, and such changes will be retained during optimization across epochs. 
Otherwise, if the information disturbs the correct prediction, the instant influence of this step will be counteracted by other steps after epochs. 
Therefore, the SC contrastive loss across epochs assists the model in dynamically extracting valuable contextual information that enhances the prediction while get rid of the noises.

Denoting the citance for sample $i$ as $s^0_i$, 
the input of \modelname~ is an enlarged context $T_i=\langle s^{-l}_i, ...,s^{-1}_i , s^0_i, s^1_i,..., s^{l}_i\rangle$, representing a sequence of $2l+1$ sentences where $l$ is a hyperparameter defining the maximum one-side range.
SC produces contexts with perturbed ranges, 
$$\mathcal{SC}(T_i) = \{\langle s^{-b}_i,...,s^0_i, ...,s^v_i \rangle|\forall b,v, -l\leq -b\leq 0\leq v\leq l\}/\{T_i\}.$$
Each transformed sample is comprised of the citance, $b$ preceding sentences, and $v$ succeeding sentences.
SC can produce at most $(l+1)^2-1$ transformed samples and the minimum resultant context is $\langle s_i^0 \rangle$.
Among them, we repeatedly iterate to obtain the positive pair for the current epoch $e$, $\tilde{T}^{SC}_{i,e}\sim \mathcal{SC}(T_i)$. 
With temperature hyperparameter $\tau_1$, the contrastive loss is 
\begin{equation}
L^{SC}=-\frac{1}{|\mathcal{B}|}\sum_{i\in \mathcal{B}}\log \frac{\exp(z_i^\mathsf{T}\tilde{z}_{i,e}^{SC}/\tau_1)}{\sum_{j\in \mathcal{B}} \exp(z_i^\mathsf{T}\tilde{z}_{j,e}^{SC}/\tau_1)}.
\end{equation}

\subsection{Keyphrase Perturbation}
The intuition of the keyphrase perturbation strategy is that modifying the scientific keyphrases in the context usually does not affect the citation intention. The scientific keyphrases are associated with detailed topics, while the residue of the sentence organizes the knowledge and outlines the writing logic, playing distinct roles in scientific writing.
Hence, the residue is usually more dominant in deciding the citation intention, an example is as follows. 
\begin{example}
Here are two texts, \textbf{S1}: "The first work to do this with \underline{topic models} is [1].", 
and \textbf{S2}: "We use \underline{topic models} [1] to find hidden semantic structures in documents."
They are similar if we consider general semantics because they are both written around the technique "topic models".
However, for citation classification, the label of \textbf{S1} is "BACKGROUND" while the label of \textbf{S2} is "USE", since they express totally distinct logic regarding the intention of the target citation.
Further, consider \textbf{S3}: "The first work to do this with \underline{data augmentation} is [1].", 
which is an artificial text created by altering the technique in \textbf{S1}. 
\textbf{S3} is quite plausible to appear in a paper on the topic "data augmentation" and is very likely to belong to the same citation label "BACKGROUND" as \textbf{S1}. 
\end{example}

As illustrated in Figure~\ref{fig:framework} (c), KP recognizes and changes the scientific keyphrases within the original context to construct its positive pairs with the same residue. Through subsequent contrastive learning, this encourages the representation to model the logical semantics of sentence residue instead of specific keyphrase semantics.

We first standardize the definition of keyphrases in our research as the scientific typed keyphrases (STKs)~\cite{lahiri2024few}, which are the entities that have indispensable semantic meaning in the scientific domain, including specific proper nouns such as "BERT" and significant words or phrases in the text, such as "citation classification". Analogized to the named entities~\cite{petasis2000automatic} in the general domain which can be typed as person, organization, and location, STKs also come with types, such as task, technique, materials, and concept. In this work, we will use the terms keyphrase and STKs interchangeably. There are several tools and methods for STK extraction under the supervised setting or few-shot setting~\cite{lahiri2024few}. 
Here we consider the simple one-shot setting so as to reduce manual efforts to label for this intermediate task, and leverage the instructional generation to extract the STKs with LLMs. This training-free approach performs well even under the one-shot setting, thanks to the great generalization ability of LLMs.

Formally, for the training set $\mathcal{D}$, let $K_i$ be the STKs appeared in $T_i$ and $\mathcal{K}=\bigcup_{i\in \mathcal{D}} K_i$ be the set of observed STKs.
KP perturbs each $k\in K_i$ at a predefined probability $\beta$. We consider the following three different operations to perturb $k$, which will be invoked periodically at different epochs.
\begin{itemize}
\item \textbf{G}lobal \textbf{r}eplacement. 
The mention of $k$ is replaced by another $k'\in \mathcal{K}/\{k\}$. We additionally require $k$ and $k'$ to have the same type because different types of STKs tend to function differently in the context.

\item \textbf{L}ocal \textbf{r}eplacement.
The mention of $k$ is replaced by $k'\in K_i/\{k\}$ of the same type. It is a localized version of \textbf{Gr} that only allows replacement between keyphrases occurring in the same context, ensuring $k'$ and $k'$ are semantically relevant to each other.
    
\item \textbf{Ab}straction. 
The mention of $k$ is replaced by its type name, such as "Task-1" and "Technique-2". We add the extra numerical IDs to distinguish between perturbed keyphrases of the same type. 
This operation masks out $k$ by making it anonymous without introducing new keyphrases. 
\end{itemize}

After the perturbing STKs, KP also performs synonym replacement~\cite{wei2019eda} on the residue of the context to introduce semantic diversity in the general domain. Particularly, each word except for the stop words is perturbed at predefined probability $\gamma$ and we use the WordNet~\cite{miller1995wordnet} synonym base $\mathcal{SN}$ to query for the synonyms. 
This step introduces general semantic differences between the positive pairs, in case there are few or no STKs in the context.
Let the scheduled perturbation operation for epoch $e$ is $\mathsf{Op}_e\in$ \{\textbf{Gr}, \textbf{Lr}, \textbf{Ab}\}, the overall KP algorithm to generate $\tilde{T}^{KP}_{i,e}$ is in 
Algorithm~\ref{algo:kp}.
\begin{algorithm}[t]
    \renewcommand{\algorithmicrequire}{\textbf{Input:}}
    \renewcommand{\algorithmicensure}{\textbf{Output:}}
    \caption{Keyphrase perturbation for epoch $e$}
    \label{algo:kp}
    \begin{algorithmic}[1]
        \Require  $K_i$, $T_i$, $\mathsf{Op}_e$, $\beta, \gamma$, synonym base $\mathcal{SN}$, global STK base $\mathcal{K}$ 
        
        \State $\tilde{T}_{i,e}^{KP}\leftarrow T_i$
        \For {$k\in K_i$}
            \State Sample indicator $q\sim Bernoulli(\beta)$
    	\If {$q=1$}
                \State Perturb $k$ in $\tilde{T}_{i,e}^{KP}$ with $\mathsf{Op}_e$
            \EndIf
        \EndFor
        
        \State Split $T_i$ into word list $\mathcal{W}$ and eliminate the stop words.
        \For {$w\in \mathcal{W}$ and $w$ not in any $k\in K_i$}
            \If {$w$ has synonyms in $\mathcal{SN}$}
                \State Sample indicator $q\sim Bernoulli(\gamma)$
                \If {$q=1$}
                \State Sample $w'$ from $\mathcal{SN}[w]$ 
		        \State Replace $w$ in $\tilde{T}_{i,e}^{KP}$ with $w’$ 
                \EndIf
            \EndIf
        \EndFor
        \Ensure $\tilde{T}_{i,e}^{KP}$ 
 \end{algorithmic}
\end{algorithm}
With temperature hyperparameter $\tau_2$, the contrastive loss for epoch $e$ is 
\begin{equation}
L^{KP}=-\frac{1}{|\mathcal{B}|}\sum_{i\in \mathcal{B}}\log \frac{\exp(z_i^\mathsf{T}\tilde{z}_{i,e}^{KP}/\tau_2)}{\sum_{j\in \mathcal{B}} \exp(z_i^\mathsf{T}\tilde{z}_{j,e}^{KP}/\tau_2)}.  
\end{equation}

\subsection{Complexity Analysis}
Denote the number of trainable parameters in $\mathcal{M}$ as $N_0$, the hidden state size of $\mathcal{M}$ as $d_x=|x|$, the model size as $d_z=|z|$, the intermediate embedding size of the adapter as $d$, $C=|\mathcal{C}|$ is the size of label set, the number of trainable parameters of \modelname~ is $N_0+d(d_x+d_z)+d+d_z+C(d_z+1)$. 
With $C\ll d_z$ and $d_z<d_x$, the number of parameters is in O($N_0+2d\cdot d_x$).
Additionally, our method can work seamlessly with prevalent parameter-efficient fine-tuning methods for LLM, such as Lora, which wraps $\mathcal{M}$ and reduces $N_0$ without changing its interface with our framework.
As for time complexity, 
since SC only shortens the $T_i$ and KP performs replacement between synonyms or STKs, the length of $\tilde{T_i}$ after tokenization resembles that of $T_i$, implying similar forwarding complexity.
Assume the PLM forwarding complexity per sample is $O(|\mathcal{M}|)$, the amortized complexity to perform contrastive learning amongst the batch is $O(|\mathcal{B}|\cdot d_z^2)$, hence the training complexity of our framework per sample is $O(3|\mathcal{M}|+6d\cdot d_x+2|\mathcal{B}|\cdot d_z^2)\approx O(3|\mathcal{M}|) $ and the inference complexity is $O(|\mathcal{M}|+2d\cdot d_x)\approx O(|\mathcal{M}|)$. 
Here the time for for SC and KP can be neglected, because with $l'$ be the length of $T_i$, it takes $O(l')$ time to scan the context and generate a transformed sample, while $|\mathcal{M}|$ is generally in O($l'^2$).
\section{Experiments} 
We conduct experiments to investigate the proposed framework and answer the following research questions.
\begin{itemize}
    \item \textbf{RQ1}: How is the overall performance of our framework?
\item \textbf{RQ2}: How effective is the proposed SC strategy in extracting contextual information and defending the irrelevant noise?
\item \textbf{RQ3}: How effective and robust is the proposed KP strategy?
\end{itemize}

\subsection{Setup}
\subsubsection{Datasets}
\begin{table*}[t]
\centering
\caption{Dataset statistics.}
\begin{tabular}{l|cccc|cccccc}
\toprule
        & \multicolumn{3}{c}{Splits} & \multicolumn{6}{c}{Citation Types (\%)}\\
Dataset &\#All & \#Train & \#Validation & \#Test & Background& Compare/Contrast & Uses & Motivation &Extend & Future \\
\midrule
ACL-ARC & 1,929 &1,399 & 246& 284 &51.3&18.1&3.7&3.6 & 4.6 & 18.7\\
FOCAL &4,166 & 2,617 & 660 &889& 41.2 & 24.7 & 0.3 & 0.9 & 5.8 & 27.1 \\
ACT2 &3,000 &2,550 &450 &1,000 &54.8 & 12.2 & 5.8 &1.9 & 9.6 &15.8\\
\bottomrule
\end{tabular}
\label{tab:data}
\end{table*}
We use two domain-specific datasets and a multidisciplinary dataset, 
and each dataset consists of 6 categories as labels. 
For ACL-ARC and ACT2, we use the original test split, and reserve 15\% of the training data as the validation split since the release does not include a validation split. For FOCAL, we use the original split. The statistics are summarized in Table~\ref{tab:data}. 
\begin{itemize}
\item \textbf{ACL-ARC}~\cite{kunnath2022dynamic} is in the domain of computational linguistics, which is initially annotated and released by Jurgens et al.~\cite{jurgens2018measuring} and processed for citation classification by Cohan et al.~\cite{cohan2019structural}. 
Although the original version is used by several later works, it is pointed out that there are duplicates, data leakage, and incomplete sentences, which may be caused by limited OCR techniques in the early years. 
We use the cleaned version by Kunnath et al.~\cite{kunnath2022dynamic}.

\item \textbf{FOCAL}~\cite{grezes2023function} is from the astrophysical domain. The original labels seem to further divide the 'Compare/Contrast' into 3 fine-grained classes by sentiment (similarities, differences, neutrality). We regard them as one class to align with other datasets.

\item \textbf{ACT2}~\cite{kunnath2022act2, pride2020authoritative} is a highly heterogeneous multidisciplinary dataset that is challenging for existing citation classification methods, comprised of samples from over 20 domains including medicine, psychology, computer science, business, economics, etc. 

\end{itemize}

\subsubsection{Backbones} 
For encoder-based PLMs, we experiment with the SciBERT~\cite{beltagy2019scibert}, a bert-based model pretrained on papers from the corpus of Semantic Scholar~\footnote{www.semanticscholar.org} and owns vocabulary that is built to best match the training corpus. SciBERT results in state-of-the-art performance on a wide range of scientific domain NLP tasks and is highlighted by the previous works in citation classification~\cite{lei2024meta, kunnath2023prompting, maheshwari2021scibert}.
Following the experimental results of previous work~\cite{kunnath2023prompting}, we use a null template~\cite{logan2022cutting} that does not include any task-specific patterns. 
\begin{quote}
    \textbf{P1}: \{$T$\}. [MASK].
\end{quote}
For decoder-based LLMs, we experiment with the instruction-tuned Llama3-8B since it is one of the leading open-source LLMs among models of similar size. Specifically, we use the bfloat16 version of instruction-tuned Llama3-8B~\footnote{huggingface.co/meta-llama/Meta-Llama-3-8B-Instruct} and further apply the Low-Rank Adaptation~\cite{yu2023low} on $\mathcal{M}$ to reduce trainable parameters.  
As for the prompt, we write a task description suggested by previous work~\cite{lei2024meta} to elevate the quality of task-specific output from the LLM. The overall prompt is as follows. 
\begin{quote}
\textbf{P2}: \textit{You are provided a context from a paper P citing a paper Q, with the specific citation marked as the "\#CITATION\_TAG" tag. Please analyze the citation function of the context which represents the author’s motive or purpose for citing Q. 
Here is the context:“}\{$T$\}\textit{". Only output one word as the answer:}
\end{quote} 
Our motivation for experiments with the LLMs is to offer a potential way to take advantage of the LLMs for citation classification, as well as to shed light on the possible performance.
We did not opt for the larger models because the task is inherently in data shortage, which may not be affordable to blindly upgrade the model scales. For instance, we found finetuning 21M to 42M parameters out of the 8B is already a sweet point with maximal performance.

\subsubsection{Baselines}
We first introduce the best-performed feature-based baseline that does not include finetuning of any PLMs. (1)~\textbf{Scaffold}~\cite{cohan2019structural}: It concatenates the Glove and ElMo embeddings of the words in the context as features and employs a BiLSTM-Attention model to aggregate among them. It further designs two auxiliary tasks, predicting whether a sentence contains a citation and predicting the section name, to handle the data scarcity.

We then introduce 3 baselines that are dedicated to finetune encoder-based PLM. (2)~\textbf{TRL}~\cite{shui2024fine}: It is a multi-task learning framework that uses the labeled data from auxiliary datasets to aid the fine-tuning on the primary dataset. A task relation learning procedure automatically computes the task weights. In our experiments, we use ACL-ARC and FOCAL as the auxiliary datasets for each other; for ACT2, neither of the other datasets is helpful so we only use itself.   
(3)~\textbf{IREL}~\cite{maheshwari2021scibert}: It is the winning system in the 2021 SDP Citation Context Classification Shared task~\cite{kunnath2021overview}, which fine-tunes SciBERT and a linear classifier end-to-end with a class-balanced classification loss~\cite{king2001logistic}.
(4)~\textbf{PET}~\cite{kunnath2023prompting}: It explored several closed-form prompts to fine-tune SciBERT in the Pattern Explicit Tuning style~\cite{ahmad2017pattern}. It reports \textbf{P1} as one of the best prompts across different datasets, resulting in previously state-of-the-art performance. 

For the decoder-based LLMs, we implement 2 straightforward baselines as there are no previous works dedicated to tuning them for citation classification. 
(5)~\textbf{IFP}~\cite{kunnath2023prompting}: It is a training-free method that adopts the model for text classification via instruction-following prompting and searches for the textual label from the decoded response. The prompt is listed in the Appendix.
(6)~\textbf{LoRA}~\cite{yu2023low}: It is a PEFT technique introducing a small number of trainable rank decomposition matrices to adapt the pretrained LLMs. We use \textbf{P2} in this baseline. 

\begin{table*}[t]
\centering
\caption{Overall performance comparison. The best score is bolded and the second-best score is underlined for each backbone. And * denotes significant improvements (measured by t-test, p<0.05) compared with other baselines with the same backbone.}
\begin{tabular}{l|l|cccccc}
\toprule
&&\multicolumn{2}{c}{ACL-ARC}&\multicolumn{2}{c}{FOCAL} & \multicolumn{2}{c}{ACT2} \\
Method & Backbone & Macro-f1 & Accuracy & Macro-f1 & Accuracy & Macro-f1 & Accuracy \\
\midrule
Scaffold & NA & $0.496\pm0.021$ &$0.649\pm 0.012$ & $0.145\pm 0.041$& $0.447\pm0.133$ & 0.146$\pm$0.006 & 0.363$\pm$0.017 \\
\midrule
IREL &SciBERT &$0.614\pm 0.037$ & $0.721\pm0.025$ & $0.580\pm0.086$ & $0.742 \pm 0.007$ & \textbf{0.262}$\pm$0.012 & 0.468$\pm$0.028\\
TRL & SciBERT& $0.476\pm0.024$& $0.610\pm0.007$ & $0.604\pm0.009$ & \underline{0.756} $\pm$ 0.009 &0.118$\pm$0.001 & \underline{0.544}$\pm$0.000\\
PET &SciBERT & \underline{0.616} $\pm$ 0.022 & \underline{0.723} $\pm$ 0.019 & \underline{0.641} $\pm$ 0.044 &  $0.750\pm 0.008$ & \underline{0.258}$\pm$ 0.018& 0.537$\pm$0.020\\
\modelname & SciBERT & \textbf{0.665}* $\pm$ 0.018 & \textbf{0.743} $\pm$ 0.006 & \textbf{0.679} $\pm$ 0.023 & \textbf{0.777}* $\pm$ 0.005 &  0.254$\pm$0.012 & \textbf{0.563}*$\pm$0.009\\

\midrule
IFP & Llama3-8B& $0.422$ & $0.575$ & $0.243$ &  $0.398$ & 0.213 & 0.446 \\
LoRA & Llama3-8B-bfloat16& \underline{0.670} $\pm$ 0.050 & \underline{0.745} $\pm$ 0.028 &  \underline{0.670} $\pm$ 0.035 & \underline{0.757}$\pm$0.001 & \underline{0.242} $\pm$ 0.034 & \underline{0.529} $\pm$ 0.014\\ 
\modelname~ + LoRA & Llama3-8B-bfloat16 & \textbf{0.744} $\pm$ 0.010 & \textbf{0.819}* $\pm$ 0.007 
&\textbf{0.682} $\pm$ 0.024 & \textbf{0.768}* $\pm$ 0.001 &\textbf{0.266} $\pm$ 0.006	& \textbf{0.549} $\pm$ 0.017
\\
\midrule
IFP & Llama3-70B & $0.569$ & $0.701$ & $0.430$ &  $0.623$ & 0.242 & 0.545\\
\bottomrule
\end{tabular}
\label{tab:acc}
\end{table*}

\subsubsection{Implementations}
We implement our framework and all LLM baselines with the python transformer library. For other baselines, we use their authorized implementation.
We use a Llama3-70B model for the STKs extraction and the detials are in the Appendix.
%
In all experiments, the context range $l$=3, the weight decay coefficient $\omega$=0.01, the learning rate is $2e^{-5}$, the synoynym replacement ratio $\gamma$=0.1. 
In the LoRA component, $r$=16 for ACL-ARC and FOCAL, $r=8$ for ACT2, and $\alpha$=16 for all settings.
For ACL-ARC, the final hyperparamers are $d$=1024, $d_z$=256, $\tau_1$=1, $\tau_2$=1, $|\mathcal{B}|$=4 for both backbones; $\lambda_1$=0.2, 0.1, $\lambda_2$=0.1, 0.2, $\beta$=0.6, 0.4, for SciBERT and Llama3-8B. 
For FOCAL, the final hyperparamers are $\lambda_1$=0.2, $\lambda_2$=0.1, $\tau_1$=5, $\tau_2$=1 for both backbones; $d$=256, 1024, $d_z$=128, 256, $|\mathcal{B}|$=16, 4, $\beta$=0.6, 0.7, for SciBERT and Llama3-8B. 
For ACT2, the final hyperparamers are $\lambda_1$=0.1, $\tau_1$=0.1; $\lambda_2$=0.2, $\tau_2$=10 for both backbones; $d$=256, 128, $d_z$=128, 64, $|\mathcal{B}|$=16, 4, $\beta$=0.3, 0.4 for SciBERT and Llama3-8B.
The reported performance under each setting is averaged over 3 runs. 
At each run, the model is trained for at most 10 epochs with early stopping based on the summation of Macro-f1 and Accuracy computed based on the validation set.
All the experiments are conducted on a server equipped with Intel(R) Xeon(R) Gold 6240 CPU and two NVIDIA A800 (80GB Memory). 
More implementation details are summarized in the Appendix.

\subsection{RQ1: Overall Comparison}
\subsubsection{Classification Performance}
The overall comparison is reported in Table~\ref{tab:acc}. Macro-f1 represents the average performance on each class since the citation types are unevenly distributed as in Table~\ref{tab:data}. Based on the results, we have the following four main observations.
(1)~\modelname~ achieves the state-of-the-art performance, outperforming existing methods on 5 out of 6 metrics when using SciBERT and on all metrics when using the Llama3-8B backbone. 
This highlights the versatility and effectiveness of our approach with both encoder-based PLMs and decoder-based LLMs. 
On ACT2, it is hard to achieve high scores on both metrics due to the difficulties in predicting its minority classes. IREL prioritizes Macro-F1 by weighting minority classes more heavily, compared to which our framework strikes a better balance and achieves significantly higher accuracy than IREL without substantial loss in Macro-F1.
(2)~Turning to the experiments with SciBERT, PET achieves the second-best overall performance. The success of both our method and PET, which leverage prompting strategies, highlights the important role of task-specific patterns in adapting PLMs for citation classification. 
TRL attains the second-best accuracy on FOCAL and ACT2, albeit for different reasons. On FOCAL, its performance validates the effectiveness of supplementing labeled data from the auxiliary ACL-ARC dataset. However, it becomes a dummy classifier on ACT2 that overwhelmingly predicts the majority class, resulting in the worst Macro-F1 score.
(3)~In experiments utilizing decoder-based LLMs, the fine-tuning methods essentially outperform the IFP. This superiority extends even when comparing fine-tuning to IFP using the significantly larger Instruction-tuned 70B model. 
This reflects that there is still a large gap between the generation approach, which relies solely on task descriptions and examples, and the data-driven fine-tuning approaches of cutting-edge LLMs for the complex text classification task of citation classification. 
(4)~With \modelname, Llama3-8B outperforms SciBERT on ACL-ARC, achieving significantly better results. While the performance gap between Llama3-8B and SciBERT narrows on Focal and ACT2, this suggests that Focal and ACT2 may benefit more from scientific text-specialized pretraining of SciBERT. ACL-ARC appears to leverage Llama3-8B's larger scale pretraining on general domain knowledge.

\begin{table*}[t]
\centering
\caption{Ablation Study by disabling the contrastive learning loss term in \modelname. Imp. (\%) shows the average improvements of both metrics. For Llama3-8B, we use the same LoRA setting and bfloat16 version.}
\begin{tabular}{l|ccccccccc}
\toprule
&\multicolumn{3}{c}{ACL-ARC}&\multicolumn{3}{c}{FOCAL} & \multicolumn{3}{c}{ACT2} \\
SciBERT & Macro-f1 & Accuracy & Imp.(\%) & Macro-f1 & Accuracy & Imp.(\%) & Macro-f1 & Accuracy & Imp.(\%) \\
\midrule
$\lambda_1,\lambda_2=0$ &$0.616\pm0.012$ & $0.714\pm0.022$ & - & $0.641\pm0.029$ & $0.742\pm0.004$ & - &
$0.262\pm0.016$ & $0.539\pm 0.021$ & - \\
$\lambda_2=0$ &$0.660\pm0.025$ & $0.745\pm0.032$ & $5.7$ & $0.666\pm0.029$ & $0.762\pm0.006$ & $3.3$ & $0.246\pm0.036$ & $0.558\pm0.008$ & $0.3$ \\
$\lambda_1=0$ &$0.646\pm0.032$ & $0.730\pm0.017$ & $3.5$ & $0.654\pm0.004$ & $0.764\pm0.009$ & $2.5$ & $0.260\pm0.003$ & $0.549\pm0.017$ & $0.9$ \\
\midrule
Llama3-8B & Macro-f1 & Accuracy & Imp.(\%) & Macro-f1 & Accuracy & Imp.(\%) & Macro-f1 & Accuracy & Imp.(\%) \\
\midrule
$\lambda_1,\lambda_2=0$ &$0.729\pm0.007$ & $0.799\pm0.003$ & - & $0.673\pm0.008$ & $0.762\pm0.003$ & - &
$0.227\pm0.006$ & $0.544\pm 0.017$ & - \\
$\lambda_2=0$ &$0.736\pm0.021$ & $0.805\pm0.013$ & $0.9$ & $0.681\pm0.009$ & $0.769\pm0.004$ & $1.0$ & $0.248\pm0.016$ & $0.542\pm0.016$ & $2.5$ \\
$\lambda_1=0$ &$0.739\pm0.008$ & $0.804\pm0.011$ & $1.0$ & $0.675\pm0.005$ & $0.766\pm0.004$ & $0.4$ & $0.236\pm0.009$ & $0.556\pm0.019$ & $2.8$ \\
\bottomrule
\end{tabular}
\label{tab:ab}
\end{table*}

\subsubsection{Efficiency}
We now detail the absolute time costs under our experiment environments. 
(1) For \modelname, with SciBERT, it takes 17 to 23 minutes for training and 3 to 12 seconds for inference on the entire test set. 
With Llama3-8B and LoRA, it takes around 89, 160, and 131 minutes for training and 12, 71, and 53 seconds for inference on ACL-ARC, FOCAL, and ACT datasets, respectively.
The time for producing transformed samples with KP and SC is negligible in comparison. 
(2) Among the baselines, the feature-based Scaffold is the most efficient, requiring only a few minutes for training and seconds for inference. However, this efficiency comes at the cost of significant manual effort in data preprocessing, particularly in collecting and cleaning the section name feature. 
(3) All SciBERT-based baselines share similar inference times. As for the training, IREL and PET require similar training time, which is around 6 to 12 minutes. TRL involves augmenting the training data and determining auxiliary task weights, resulting in a comparable overall training time to \modelname~with SciBERT.
(4) For LLM baselines, while IFP is training-free, its inference speed is significantly slower due to the sequential decoding, requiring roughly 5 and 15 seconds per sample for Llama3-8B and Llama3-70B, respectively. This limits its practical applicability. LoRA fine-tuning, on the other hand, takes 40 to 92 minutes for training and 11 to 52 seconds for inference.

\begin{figure}
    \centering
    \includegraphics[width=\linewidth]{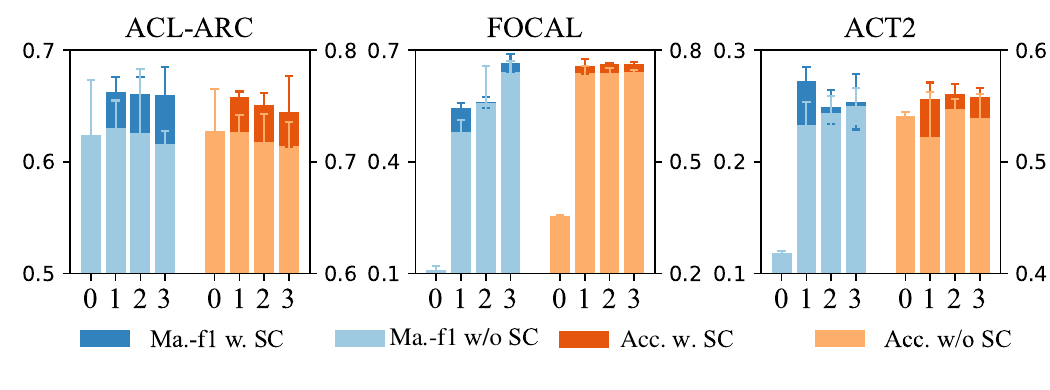}
    \caption{Ablation study of SC with different $T_i$. The x-axis is $l$, and $l=0$ corresponds to the citance.}
    \label{fig:sc}
\end{figure}
\subsection{RQ2: Investigation of Sentence-Level Cropping}
We first conduct the ablation study by disabling the contrastive learning loss term in \modelname, and list the result in Table~\ref{tab:ab}. With all datasets and backbone models, using a single strategy is better than no strategy. Combining with the overall performance in Table~\ref{tab:acc}, we can conclude that simultaneously using SC and KP further outperforms a single strategy.

Further, we adjust $l$ in $T_i=\langle s_i^{-l}, ...,s^l_i \rangle $ to input context with different ranges to the SciBERT, evaluating the performance both with and without SC in Figure~\ref{fig:sc}.
On ACL-ARC, $l=0$ is a decent input range with the highest accuracy, just as the setting of previous studies~\cite{kunnath2023prompting, maheshwari2021scibert, shui2024fine}. Increasing $l$ without SC leads to performance degradation, suggesting that the surrounding sentences introduce excessive noise that overrides beneficial information. 
For FOCAL and ACT2, expanding the context window generally improves performance. This may indicate more citation-relevant long-range dependencies and less noise compared to ACL-ARC, making a broader context more advantageous.
Across all datasets, SC consistently improves both metrics for $l=1,2,3$, and outperforms $l=0$ significantly. This validates the efficacy of SC in mitigating in-context noise and leveraging broader contexts effectively.

\begin{figure}
    \centering
    \includegraphics[width=\linewidth]{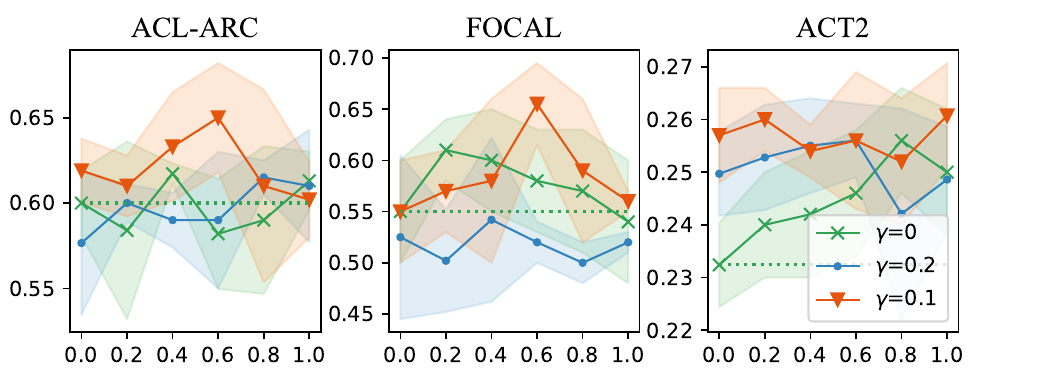}
    \caption{Performance with varying $\beta$ (x-axis) and $\gamma$. The dashed line is the performance without KP.}
    \label{fig:kp_rate}
\end{figure}
\begin{figure}
    \centering
    \includegraphics[trim=0 5 0 0, clip, width=\linewidth]{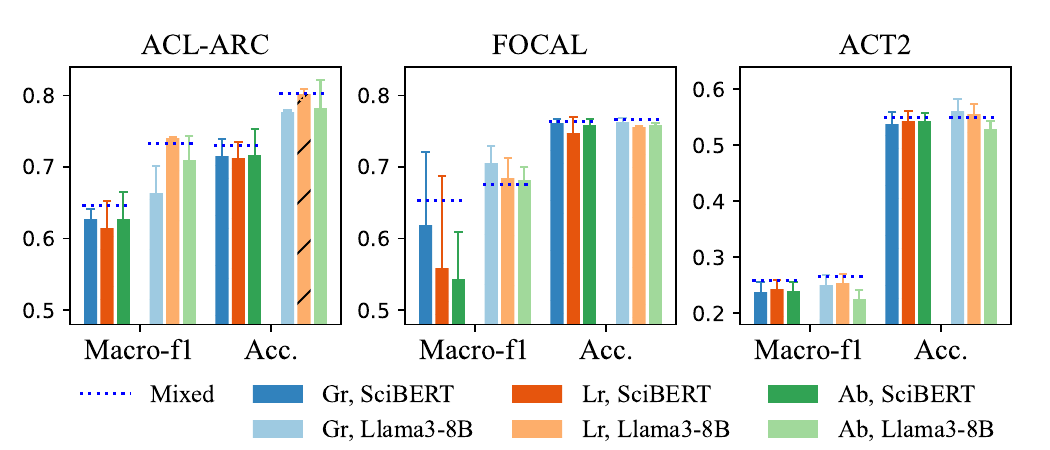}
    \caption{Performance with perturbation operation $\mathsf{Op}$.}
    \label{fig:kp_mode}
\end{figure}
\subsection{RQ3: Investigation of Keyphrase Perturbation}
We analyze the impact of the keyphrase perturbation ratio $\beta$ and synonym replacement ratio $\gamma$ on performance, reporting Macro-F1 scores with SciBERT in Figure~\ref{fig:kp_rate} (with a similar pattern for Accuracy). Increasing $\gamma$ initially improves performance within a certain range, but exceeding an optimal value leads to degradation. This suggests that introducing general domain diversity to scientific writing is beneficial to a point.
On ACL-ARC and FOCAL, performance peaks as $\beta$ increases under an optimal $\gamma$, indicating that gradually increasing the differentiation of keyphrases between positive samples enhances performance. But entirely different keywords between positive pairs may pose challenges for model optimization. On ACT2, varying $\beta$ has a less significant effect with the optimal $\gamma$, possibly due to its high heterogeneity, where keyphrases are markedly dissimilar.

We also compare different proposed perturbation operations in Figure~\ref{fig:kp_mode} by evaluating performance under single and mixed modes. In most cases, using the mixed operations achieves optimal or near-optimal results, facilitating the implementation of the framework.

\begin{figure}
    \centering
    \includegraphics[width=\linewidth]{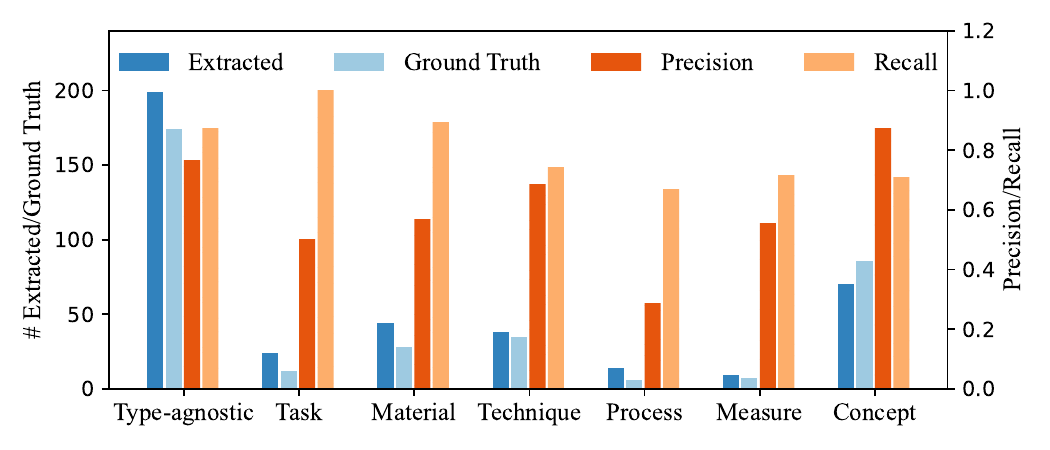}
    \caption{Extracted STKs estimated by humans.}
    \label{fig:stk_acl_arc}
\end{figure}

Finally, we analyzed the extracted STKs, beginning with a quantitative assessment. Across the ACL-ARC, FOCAL, and ACT2 datasets, we obtained 6801, 21244, and 13731 STKs, respectively. Subsequently, we conducted a qualitative analysis by randomly sampling 20 contexts from ACL-ARC. Human annotators identified 174 STKs in these samples, while the LLM extracted 199. The quality of these extractions is further detailed in Figure~\ref{fig:stk_acl_arc}. 
The main differences between the LLM-mined and manually annotated STKs reside in two aspects, neither of which poses a dangerous risk to our framework. 
First, the LLM exhibits a more lenient criterion for STK identification, resulting in a recall ranging from 0.667 to 1, which usually exceeds its precision. For example, the LLM identified "generating an initial description" as a Process, while human annotators did not acknowledge it as an STK. 
Introducing perturbations based on such high-order semantic elements, which are not strictly STKs, is unlikely to severely compromise the validity of the KP algorithm. Indeed, exploring alternative transformation strategies based on perturbing these higher-order elements, or even paraphrasing the context entirely, presents a promising avenue for future research.
Second, the LLM demonstrated a tendency to categorize a larger proportion of STKs under the "Concept" category, which encompasses STKs not readily classified into other defined types. In contrast, human annotators were more adept at identifying specific STK types. Fortunately, the KP algorithm can perturb the STK in the original context regardless of its assigned type. The type-agnostic precision and recall, at 76 and 87\%, are high, lending strong support to our KP strategy.
\section{Conclusion}
We introduce a framework, \modelname, to fine-tune PLMs for citation classification via self-supervised contrastive learning.
Our framework employs sentence-level cropping and keyphrase perturbation strategies to construct contrastive pairs without the need for labels. 
Taking the citation context as input, our framework acquires task-specific representation from the output of PLMs, and performs contrastive learning together with the supervised loss over the labeled data. 
Experiments with three benchmark datasets demonstrated the superiority of our framework.


\begin{acks}
Lei Chen’s work is partially supported by National Key Research and Development Program of China Grant No. 2023YFF0725100, National Science Foundation of China (NSFC) under Grant No. U22B2060, Guangdong-Hong Kong Technology Innovation Joint Funding Scheme Project No. 2024A0505040012, the Hong Kong RGC GRF Project 16213620, RIF Project R6020-19, AOE Project AoE/E-603/18, Theme-based project TRS T41-603/20R, CRF Project C2004-21G, Guangdong Province Science and Technology Plan Project 2023A0505030011, Guangzhou municipality big data intelligence key lab, 2023A03J0012, Hong Kong ITC ITF grants MHX/078/21 and PRP/004/22FX, Zhujiang scholar program 2021JC02X170, Microsoft Research Asia Collaborative Research Grant, HKUST-Webank joint research lab and 2023 HKUST Shenzhen-Hong Kong Collaborative Innovation Institute Green Sustainability Special Fund, from Shui On Xintiandi and the InnoSpace GBA.
Yongqi Zhang’s work is supported by Guangdong Basic and Applied Basic Research Foundation  2025A1515010304, and Guangzhou Science and Technology Planning Project 2025A03J4491.
\end{acks}

\bibliographystyle{ACM-Reference-Format}
\balance
\bibliography{33-Citss}

\appendix

\begin{table*}
\centering
\caption{STKs statistics. "Type-agnostic" is the total quantity of STKs. Under each class, we report the average number of keyphrases of that class per sample.}
\begin{tabular}{l|c|cccccc}
\toprule
Dataset & Type-agnostic & Task & Material & Technique & Process & Measure & Concept \\
\midrule
ACL-ARC & 6,801 &1.14&1.87&1.72&0.76&0.48&3.07 \\
FOCAL &21,244 & 0.91 & 3.73 &1.44 &1.42 &2.48 & 4.42 \\
ACT2 &13,731 & 1.11 & 2.20 & 0.83 & 0.73 & 1.03 & 4.15 \\
\bottomrule
\end{tabular}
\label{tab:stks}
\end{table*}

\section{Hyperparameter Ranges}
We tune $\lambda_1, \lambda_2 \in [0.01, 0.3]$, $\tau_1, \tau_2\in [0.1, 20]$, the model dimension $d_z\in\{256, 128,64\}$, $d\in \{1024, 512, 256, 128\}$, the batch size $|\mathcal{B}|\in\{4,8,16\}$, learning rate $lr\in\{ie^{-5}\}_{i\in\{1,2,4,10\}}$,
dropout rate $dr=\{ie^{-2}\}_{i\in\{0,1,5\}}$. 
For the Llama3 experiments, we tune the lora rank $r\in\{8,16,32, 64\}$,  the lora alpha $\alpha\in\{8,16,32\}$.
\section{Details of STKs}
The statistics of the extracted STKs are summarized in Table~\ref{tab:stks}.

\begin{figure}
    \centering
    \includegraphics[width=\linewidth]{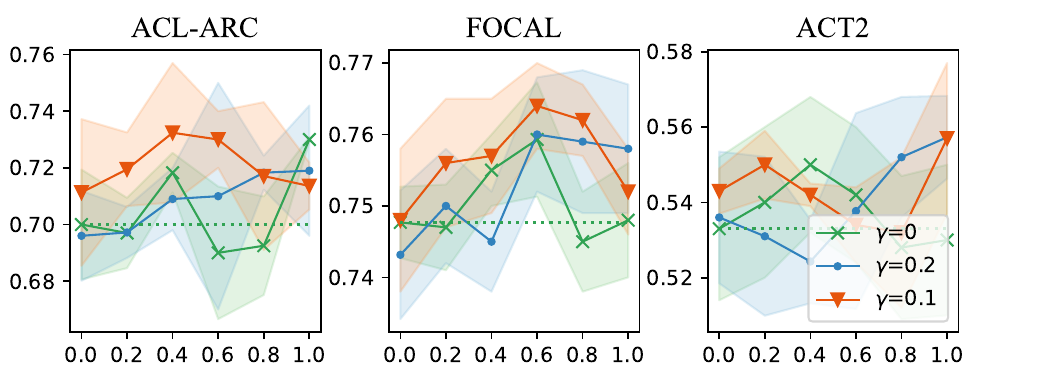}
    \caption{Accuracy with varying $\beta$ (x-axis) and $\gamma$. The dashed line is the performance without KP.}
    \label{fig:kp_rate_mi}
\end{figure}
\section{Supplementary Experiments}
Figure~\ref{fig:kp_rate_mi} is the respective Accuracy of Figure~\ref{fig:kp_rate}. 

\section{Instructions for LLMs}

\begin{tcolorbox}[title = {Prompt for STKs extraction}]
You are provided a context from a paper P, and please ignore the \#CITATION\_TAG. Your task is to identify scientific keyphrases from the context. Each scientific keyphrase belongs to one of the following classes:\\

\noindent - [Task]: The scientific problem or research focus addressed in the paper. It outlines the specific objectives or questions that the study aims to answer. 

\noindent - [Material]: All materials utilized in the study, such as experimental tools, datasets, and the objects or subjects of investigation. It details the resources of the research.

\noindent - [Technique]: The specific methods, models, frameworks, or systems. It identifies the approaches taken to analyze data or solve problems.

\noindent - [Process]: It describes a sequence of steps or operations involved in a particular procedure, algorithm, or workflow. It emphasizes the procedural aspects.

\noindent - [Measure]: This class pertains to the metrics, indicators, or criteria used to assess or quantify the outcomes of the study.

\noindent - [Concept]: This category encompasses scientific 
keyphrases that do not fit into the aforementioned classes. It may include phenomena, theoretical terms, or entities relevant to the field of study.\\

\noindent Output your answer only in JSON format and be consistent with the text in the original context. Specifically, if there is any keyphrase of a certain class, use the class label as the key and the list of keyphrases as the value. \\

\noindent Here is an example: "The framework represents a generalization of several predecessor NLG systems based on Meaning-Text Theory: FoG (Kittredge and Polgu~re, 1991), LFS (Iordanskaja et al., 1992), and JOYCE (Rambow and Korelsky, 1992). The framework was originally developed for the realization of deep-syntactic structures in NLG ( \#CITATION\_TAG )"

\noindent Output:\{'Technique': ['NLG systems', 'FoG', 'LFS', 'JOYCE', 'Meaning-Text Theory'], 'Concept':['deep-syntactic structures']\}

\noindent Here is the context:\{$T$\}
\end{tcolorbox}

\begin{tcolorbox}[title = {Prompt for the IFP baseline}]
You are provided a context from a paper P citing a paper Q, with the specific citation marked as the '\#CITATION\_TAG' tag. Please analyze the citation function of the context, which represents the author’s motive or purpose for citing Q. The six classes of citation functions are: \\

\noindent - [BACKGROUND]: The cited paper Q provides relevant information or is part of the body of literature in this domain.

\noindent - [COMPARES\_CONTRASTS]: The citing paper P expresses similarities or differences to, or disagrees with, the cited paper Q.

\noindent - [EXTENSION]: The citing paper P extends the data, methods, etc. of the cited paper Q.

\noindent - [FUTURE]: The cited paper Q is a potential avenue for future work.

\noindent - [MOTIVATION]: The citing paper P is directly motivated by the cited paper Q.

\noindent - [USES]: The citing paper P uses the methodology or tools created by the cited paper Q. \\

\noindent Here is the context: "\{$T$\}"\\

\noindent Only output the most appropriate class to categorize \#CITATION\_TAG and enclose the label within square brackets [].
\end{tcolorbox}

\end{document}